\documentclass[sigconf]{acmart}
\usepackage{multirow}
\usepackage{subfig}
\usepackage{graphicx}
\usepackage{cleveref}
\usepackage[utf8]{inputenc}
\newcommand{\etal}{et al.}

\makeatletter
\newcommand\thankssymb[1]{\textsuperscript{\@fnsymbol{#1}}}
\makeatother

\AtBeginDocument{%
  \providecommand\BibTeX{{%
    \normalfont B\kern-0.5em{\scshape i\kern-0.25em b}\kern-0.8em\TeX}}}

\copyrightyear{2024}
\acmYear{2024}
\setcopyright{acmlicensed}
\acmConference[LGM3A '24]{Proceedings of the 2nd Workshop on Large Generative Models Meet Multimodal Applications }{October 28-November 1 2024}{Melbourne, VIC, Australia}
\acmBooktitle{Proceedings of the 2nd Workshop on Large Generative Models Meet Multimodal Applications (LGM3A '24), October 28-November 1 2024, Melbourne, VIC, Australia}
\acmDOI{10.1145/3688866.3689125}
\acmISBN{979-8-4007-1193-0/24/10}

\begin{document}

\title{SynthDoc: Bilingual Documents Synthesis for Visual Document Understanding}


\author{Chuanghao Ding\thankssymb{1}\thankssymb{2}}\thanks{\thankssymb{1} Equal contribution.}\thanks{\thankssymb{2} Work was done during internship at SenseTime Research.}
\affiliation{%
  \department{State Key Laboratory for Novel Software Technology}
  \institution{Nanjing University}
  \institution{SenseTime Research}
  \city{Nanjing}
  \country{China}}
\email{ch777.ding@smail.nju.edu.cn}

\author{Xuejing Liu\thankssymb{1}}
\affiliation{%
  \institution{SenseTime Research}
  \city{Shanghai}
  \country{China}}
\email{xuejing931210@gmail.com}

\author{Wei Tang\thankssymb{2}}
\affiliation{%
  \institution{SenseTime Research}
  \city{Shanghai}
  \country{China}}
\email{weitang@njust.edu.cn}

\author{Juan Li}
\affiliation{%
  \department{State Key Laboratory for Novel Software Technology}
  \institution{Nanjing University}
  \city{Nanjing}
  \country{China}}
\email{juanli@smail.nju.edu.cn}

\author{Xiaoliang Wang}
\affiliation{%
  \department{State Key Laboratory for Novel Software Technology}
  \institution{Nanjing University}
  \city{Nanjing}
  \country{China}}
\email{waxili@nju.edu.cn}

\author{Rui Zhao}
\affiliation{%
  \institution{SenseTime Research}
  \city{Shanghai}
  \country{China}}
\email{zhaorui@sensetime.com}

\author{Cam-Tu Nguyen\thankssymb{3}}\thanks{\thankssymb{3} Corresponding author}
\affiliation{%
  \department{State Key Laboratory for Novel Software Technology}
  \department{School of Artificial Intelligence}
  \institution{Nanjing University}
  \city{Nanjing}
  \country{China}}
\email{ncamtu@nju.edu.cn}

\author{Fei Tan\thankssymb{3}}
\affiliation{%
  \institution{SenseTime Research}
  \city{Shanghai}
  \country{China}}
\email{tanfei2007@gmail.com}
\renewcommand{\shortauthors}{Chuanghao Ding and Xuejing liu, et al.}

\begin{abstract}
  This paper introduces SynthDoc, a novel synthetic document generation pipeline designed to enhance Visual Document Understanding (VDU) by generating high-quality, diverse datasets that include text, images, tables, and charts. Addressing the challenges of data acquisition and the limitations of existing datasets, SynthDoc leverages publicly available corpora and advanced rendering tools to create a comprehensive and versatile dataset. Our experiments, conducted using the Donut model, demonstrate that models trained with SynthDoc's data achieve superior performance in pre-training read tasks and maintain robustness in downstream tasks, despite language inconsistencies. The release of a benchmark dataset comprising 5,000 image-text pairs not only showcases the pipeline's capabilities but also provides a valuable resource for the VDU community to advance research and development in document image recognition. This work significantly contributes to the field by offering a scalable solution to data scarcity and by validating the efficacy of end-to-end models in parsing complex, real-world documents.
\end{abstract}

\ccsdesc[500]{Applied computing~Multi / mixed media creation}

\keywords{Visual Document Understanding, End-to-End Document Parsing, Synthetic Document Generation}



\maketitle

\section{Introduction}
Visual Document Understanding (VDU) is a complex endeavor that seeks to decipher and interpret information from documents across a spectrum of formats and layouts~\cite{harley2015evaluation,jaume2019funsd,pfitzmann2022doclaynet,wang2021layoutreader, li2020docbank}. The objective of VDU is to develop algorithms capable of grasping the content, structure, and context of documents, thereby enabling tasks such as document classification~\cite{harley2015evaluation}, text detection~\cite{jaume2019funsd,park2019cord}, layout analysis~\cite{pfitzmann2022doclaynet,wang2021layoutreader}, and object detection~\cite{li2020docbank,li2019tablebank}.

Current research in VDU predominantly employs two methodologies: one~\cite{xu2020layoutlm,xu2020layoutlmv2,huang2022layoutlmv3,liao2023doctr,bai2022wukong,liu2023large} relies on OCR technology to convert document images into text for subsequent processing, while the other~\cite{appalaraju2023docformerv2,donut,lv2023kosmos,dhouib2023docparser,Pix2struct,liu-etal-2023-matcha,blecher2023nougat} adopts an end-to-end approach, analyzing the document images directly. The pre-training and fine-tuning paradigm is extensively utilized in multimodal learning~\cite{transcp, guo2024makes, donut, liu2023deeply, blecher2023nougat, Pix2struct}. The end-to-end approach leverages this paradigm to incorporate robust text recognition capabilities into the model, addressing the limitations of OCR accuracy and achieving high processing efficiency. A common pre-training task is the text reading task, and previous studies~\cite{Pix2struct,donut} have demonstrated its efficacy in enhancing model performance across various downstream tasks, such as document parsing and document Visual Question Answering (VQA). Therefore, leveraging the text reading task to bolster the capabilities of the base model is of paramount importance.

The data requirements for the text reading task encompass two main aspects: high-quality document images and corresponding text annotations that reflect the reading order. Obtaining such data is intricate, with existing methods either depending on large-scale public document datasets and additional OCR models to generate pseudo-labels~\cite{donut} or relying on complex data processing pipelines to scrape document data from the web~\cite{weber2024wordscape}. However, these methods often result in low-quality labels, face copyright restrictions, and contend with data noise. Moreover, they typically focus only on specific elements within document images, such as text or certain document components. For example, Nougat~\cite{blecher2023nougat} and KOSMOS-2.5~\cite{lv2023kosmos} concentrate on table parsing, while MatCha~\cite{liu-etal-2023-matcha} emphasizes chart rendering. It is rare to find a dataset that encompasses all document elements simultaneously. A recent approach Vary~\cite{wei2023vary}, while employing rendering of various document types, has utilized only over 10 templates, which falls short in terms of the richness of document layouts.

To tackle the limitations in document layout richness and the challenges associated with data acquisition, we introduce SynthDoc, a synthetic document generation pipeline. This pipeline is designed to create datasets that include text, images, tables, and a variety of charts. We begin by aggregating publicly available datasets, which have been validated on large language or multimodal models, to form our text and image corpora. We then enhance the TableGeneration~\cite{TableGeneration} to produce a diverse set of tables, and use tools like pandas~\cite{reback2020pandas}, Matplotlib~\cite{Hunter:2007}, and ECharts~\cite{li2018echarts} to generate chart-table pairs, thus expanding our chart data corpus. Therefore, our approach provides three distinct advantages: 1) Synthdoc can leverage redundant, open-resources NLP datasets to generate high-resolution, coherent content for multimodal model training.
2) Synthdoc is developed with high efficiency, precision, and dynamically customizes document layouts and features robust scalability.
3) The synthesized data include comprehensive content and structural annotations, facilitating the pre-training of structured document parsing models based on LLMs. Synthetic data can effectively complement the expensive manually labeled real datasets.

Our comprehensive experiments, leveraging the Donut model, have yielded compelling results that underscore the efficacy of the SynthDoc pipeline. The models trained with our synthesized document images have achieved remarkable accuracy in the pre-training read task, demonstrating a keen ability to parse both Chinese and English text, as well as tables and charts within the generated datasets. This proficiency extends to the fine-tuning phase of downstream tasks, where the models maintain a high level of performance despite the primary and secondary tasks involving languages that are not always consistent.

Furthermore, we have conducted visual analyses of the models' parsing capabilities on more complex, real-world documents. Despite the relatively limited variety of document types synthesized by our pipeline, the models have shown commendable results in parsing these intricate documents. A particularly surprising finding pertains to the chart parsing capabilities. In instances where scatter plots did not explicitly label the x-axis, our models were able to accurately infer the horizontal coordinates. This suggests that the models trained with our rendered data possess a certain level of spatial understanding and an awareness of the sequence among numerical values.

In response to the absence of comprehensive public datasets for model validation in document image parsing, we have released a set of 5,000 images based on the SynthDoc pipeline. This release not only showcases the quality and diversity of the document data we generate but also provides a benchmark for the document image recognition community to advance and develop new methodologies.

In summary, the key contributions of this paper are as follows:
\begin{itemize}
\item SynthDoc Pipeline: We introduce a novel synthetic data pipeline for document images, named SynthDoc, which utilizes publicly available text or text-image pairs along with rendered tables and charts. This pipeline is capable of simultaneously generating text, images, tables, and various types of charts within document images.

\item Benchmark Release: We have made available to the research community a benchmark dataset consisting of 5,000 image-text pairs. This release aims to highlight the robustness of the data produced by our pipeline and to support further research and development in the area of document image parsing.

\item Experimental Validation: Through experiments based on the Donut model, we have demonstrated that our proposed dataset and training methodology lead to a significant enhancement in the model's document image parsing capabilities. Additionally, the models trained with our approach maintain competitive performance across a range of downstream tasks.
\end{itemize}

\begin{figure*}[tb]
  \centering
  \includegraphics[width=1.0\textwidth]{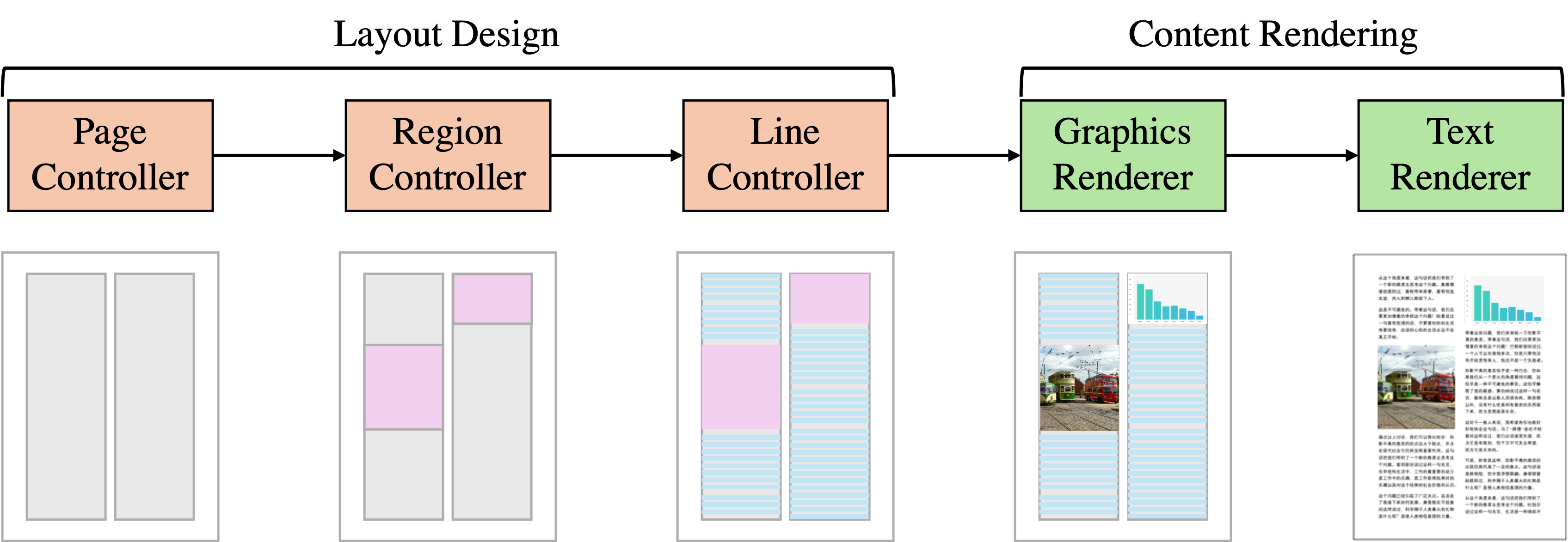}
  \caption{
    The pipeline of Document Image Synthesis, including layout design and content rendering. The layout design involves planning at three scales: full-page, regional, and line-by-line. Content rendering creates both visual graphics and textual content.
  }
  \label{fig:pipeline}
\end{figure*}

\section{Related Work}
\subsection{Image Document Data}
Deep learning-based document image understanding has consistently been recognized as a significant and intricate work, and many datasets have been proposed to parse and understand document images from different perspectives. For example, FUNSD~\cite{jaume2019funsd} is utilized for form understanding. RVL-CDIP~\cite{harley2015evaluation}is employed for document classification. PubLayNet~\cite{publaynet}is utilized for document layout analysis in our study. However, these datasets fail to meet the requirements of recent end-to-end methods, which rely on large amounts of document image data for pretraining. Some approaches~\cite{donut,davis2022end} parse existing document datasets, such as IIT-CDIP~\cite{iit-cdip}, by commercial OCR models. However the quality of datasets obtained by such methods is constrained by OCR accuracy,  and utilizing commercial OCR models can be costly. Another approachs~\cite{weber2024wordscape,lv2023kosmos,wang2021layoutreader} rely on the crawler techniques to collect extensive data from the internet, extracting document image data through parsing and filtering, which often yield datasets with considerable noise, due to the complexity of the document, and are subject to copyright restrictions. Unlike these methods, we collect existing web-scale datasets~\cite{wei2023skywork,laurenccon2024obelics,schuhmann2021laion,schuhmann2022laion,gadre2024datacomp,xue2020mt5,penedo2024refinedweb} that have been used by large language models or multimodal large language models, employing a synthetic approach to obtain document image data, which can yield clean data and include complex elements such as charts.

\subsection{Text Reading Task}
As the end-to-end multimodal model evolves, the task of text reading within document images has gained increasing attention from scholars, affirming its significant value in the field. For example, Donut~\cite{donut} is pre-trained on document images and their associated text annotations, reading text from images one by one according to previous text contexts. Nougat~\cite{blecher2023nougat} follows Donut's model and training approach, with a specialized focus on the domain of scientific papers, adeptly reading texts, tables, and formulas using markup language. DocParser~\cite{dhouib2023docparser} introduces the Masked Document Reading method, which is designed to enhance the model's reasoning capabilities by predicting the text situated within the masked regions. UReader~\cite{UReader} utilizes text reading task to train multimodal large language model, and proposes to predict text from any position of document images, which ensures the model can read different parts of texts with the context. Pix2struct~\cite{Pix2struct} found that the text reading task showed a strong curriculum learning effect, using it as warmup phase resulted in more stable training, faster convergence, and better performance. It is worth noting that all of these tasks require millions of document images, kosmos2.5~\cite{lv2023kosmos} collected 324.4M data from public datasets and web, such as IIT-CDIP~\cite{iit-cdip}, arxiv, and GitHub. However, these data are difficult to obtain and have copyright restrictions, so we propose a data rendering pipeline for text reading task to improve the model's understanding of dual-language documents.

\begin{figure*}[tb]
  \centering
    \subfloat[ ]{
      \begin{minipage}[b]{0.48\linewidth}
      \centering
      \includegraphics[height=2in]{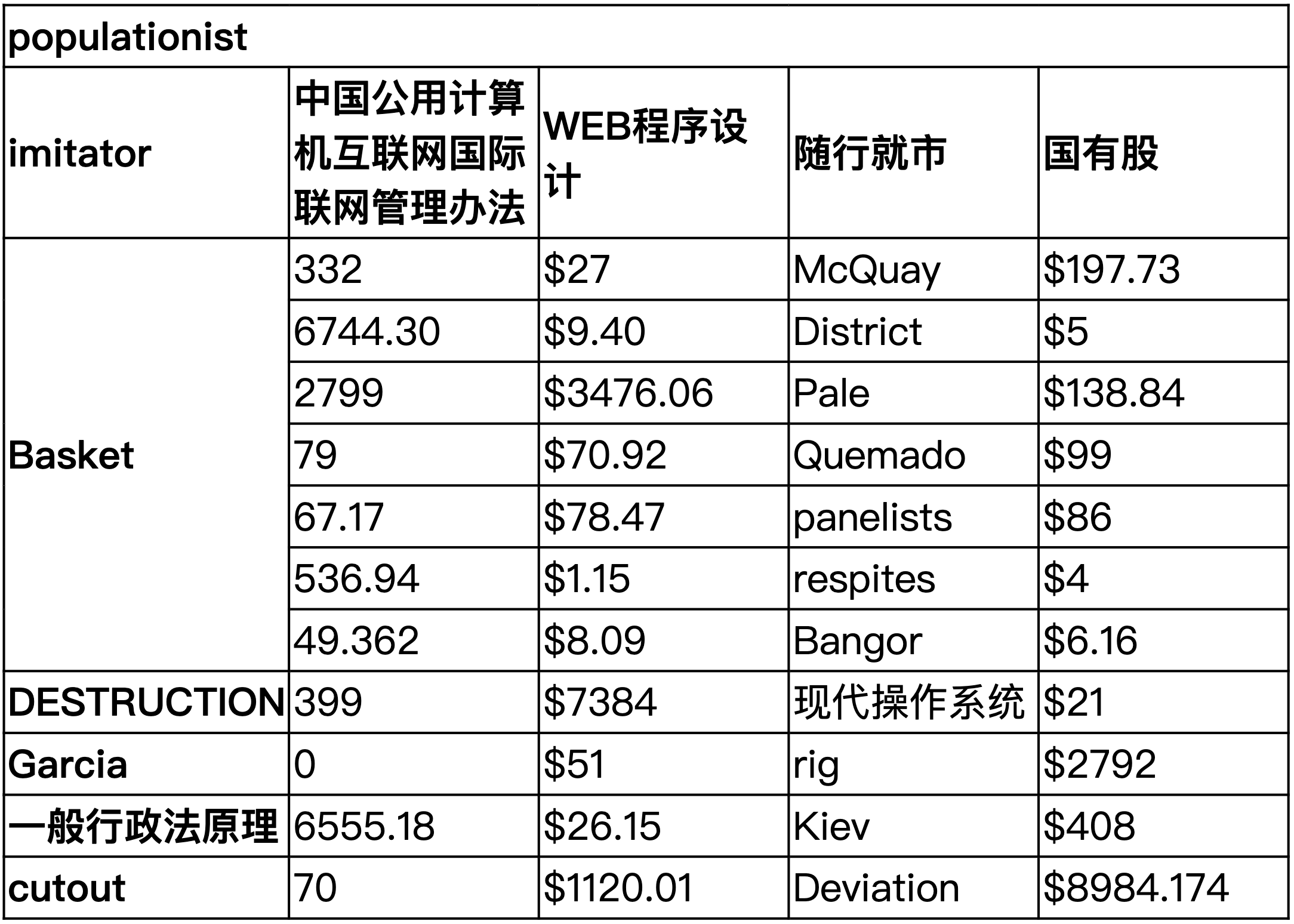}
      \label{fig:tableWboard}
      \end{minipage}
    }
    \subfloat[ ]{
      \begin{minipage}[b]{0.48\linewidth}
      \centering
      \includegraphics[height=2in]{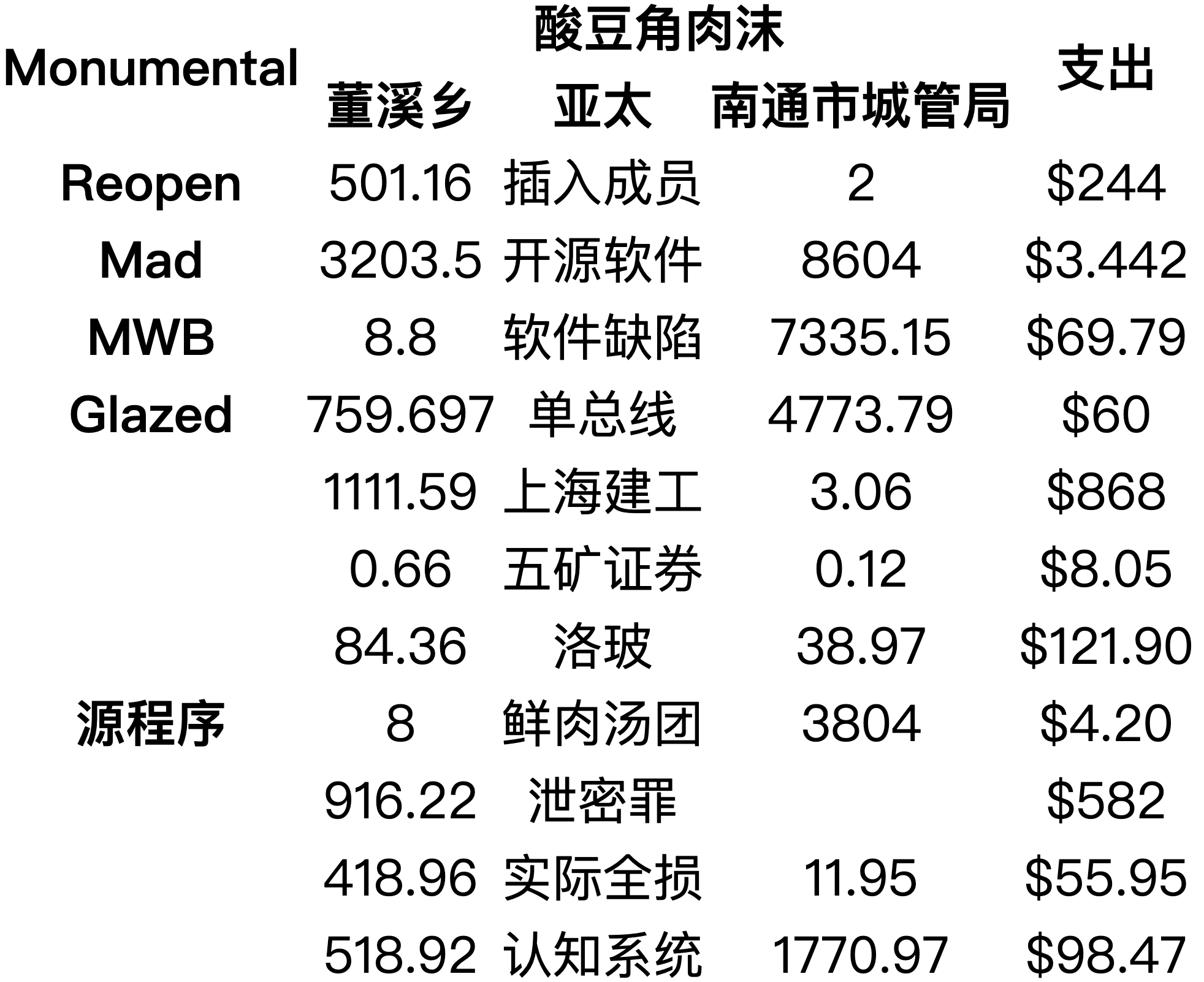}
      \label{fig:tableNboard}
      \end{minipage}
    }
  \caption{Gridlined and gridless table renderings.}
  \label{fig:table}
\end{figure*}

\subsection{Synthetic Document Image}
Document image data generation has been widely concerned in the field of visual document understanding. Some document image generation algorithms, based on GAN networks, generate plausible document images, emphasizing the diversity and quality of generated documents. For example, Biswas \etal\cite{biswas2021docsynth} utilize the GAN model to generate diverse and credible document images based on the provided layout. Zheng \etal\cite{zheng2019content-aware} proposed a layout depth generation model for graphic design, which implicitly captured the influence of visual and text content on layout, and synthesized complex layout design according to the visual and text semantics input by users. However, these methods do not consider the annotation information used for visual document understanding, the quality and size of the generated images are limited by the model, and additional training models are required for different languages, which is inefficient. Other methods generate document and ground truth pairs for specific visual document understanding tasks. For document layout analysis, Pisaneschi \etal\cite{pisaneschi2023automatic_generation} generates document layout information based on LayoutTransformer\cite{layouttransformer} and additional post-processing methods which fill in the corresponding texts, images, and Mathematical objects based on the model or the collected corpus. Ling \etal\cite{ling2021document_domain} proposed the document domain randomization approach to simulate the document layout, and then randomly fill in collected elements such as texts and images. For pretraining of Document Intelligence tasks, Biten \etal\cite{ocr-idl} generates large-scale pre-training datas with OCR annotation information on IDL datasets based on commercial OCR tools. However, the current pre-training of intelligent document understanding based on large language models relies on document image parsing tasks, and the existing data can no longer meet the training demands, so we propose a new data set generation pipeline to synthesize accurate, clear, logical and coherent document parsing datasets to adapt to the development of visual document understanding.

A similar effort to this paper is donut\cite{donut}, which uses a portion of generated data to supplement data in different languages. The difference is that their work randomly pastes text into images and ignores layout information and structured elements such as tables, charts and images.

\section{Document Image Synthesis}
In this section, we delve into the pipeline for generating document images, which is primarily composed of two key components: layout design and content rendering, as shown in Fig.~\ref{fig:pipeline}. The layout design encompasses the architectural planning at three distinct scales: the entire page, individual regions, and lines of text. This meticulous arrangement ensures that the document's structure aligns with conventional reading habits while maximizing visual diversity.
Content rendering, on the other hand, is responsible for the creation of both graphic and textual elements. This phase includes the rendering of graphics, which can consist of tables, images, and charts, as well as the rendering of text. Each element is crafted with attention to detail, ensuring that the final document image not only conveys information accurately but also presents it in an aesthetically pleasing and reader-friendly manner.

\subsection{Layout Design}
The document image synthesis pipeline comprises three integral components: the Page Controller, Region Controller, and Line Controller. The Page Controller ensures a consistent and visually appealing layout by defining and maintaining layout elements and typographical attributes. The Region Controller segments the document into distinct areas for various content types, facilitating a logical and balanced composition. Lastly, the Line Controller meticulously organizes text, applying typographical rules to enhance readability and engagement. Together, these components work to create structured, professional-looking documents that are both informative and aesthetically pleasing.

\subsubsection{Page Controller.}
The Page Controller is instrumental in establishing a consistent and visually appealing layout for single-page documents. It sets and maintains the uniformity of layout elements such as data areas, page margins, and the spacing between segments and lines. Additionally, it oversees the font size and color palette, ensuring that the document's visual presentation is coherent and reader-friendly. This component's role is critical in creating a structured and professional look that enhances the document's overall readability and impact.

\begin{figure*}[tb]
  \centering
  \subfloat[ ]{
    \begin{minipage}[b]{0.48\linewidth}
    \centering
    \includegraphics[height=2.3in]{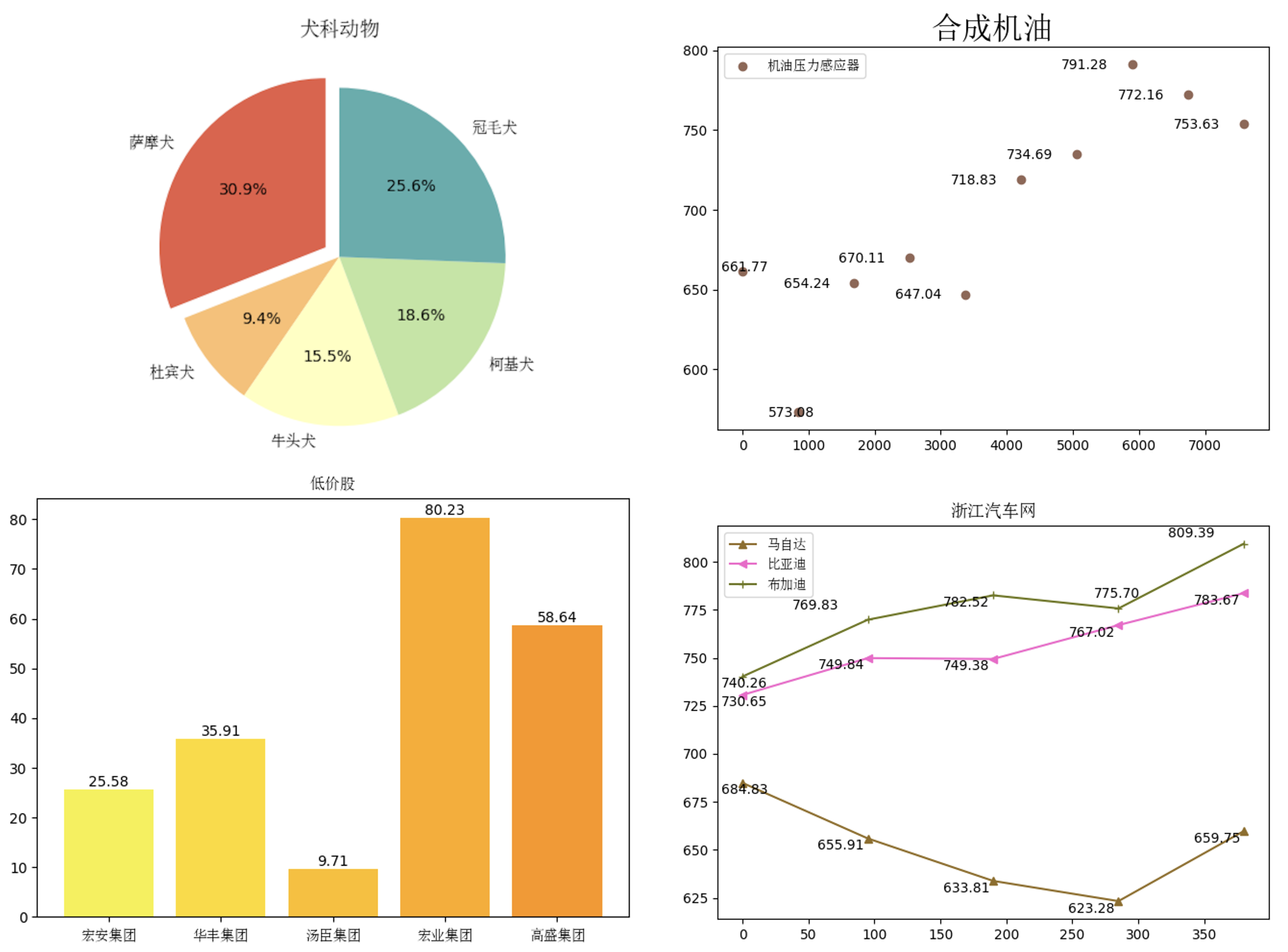}
    \label{fig:element4}
    \end{minipage}
  }
  \subfloat[ ]{
    \begin{minipage}[b]{0.48\linewidth}
    \centering
    \includegraphics[height=2.3in]{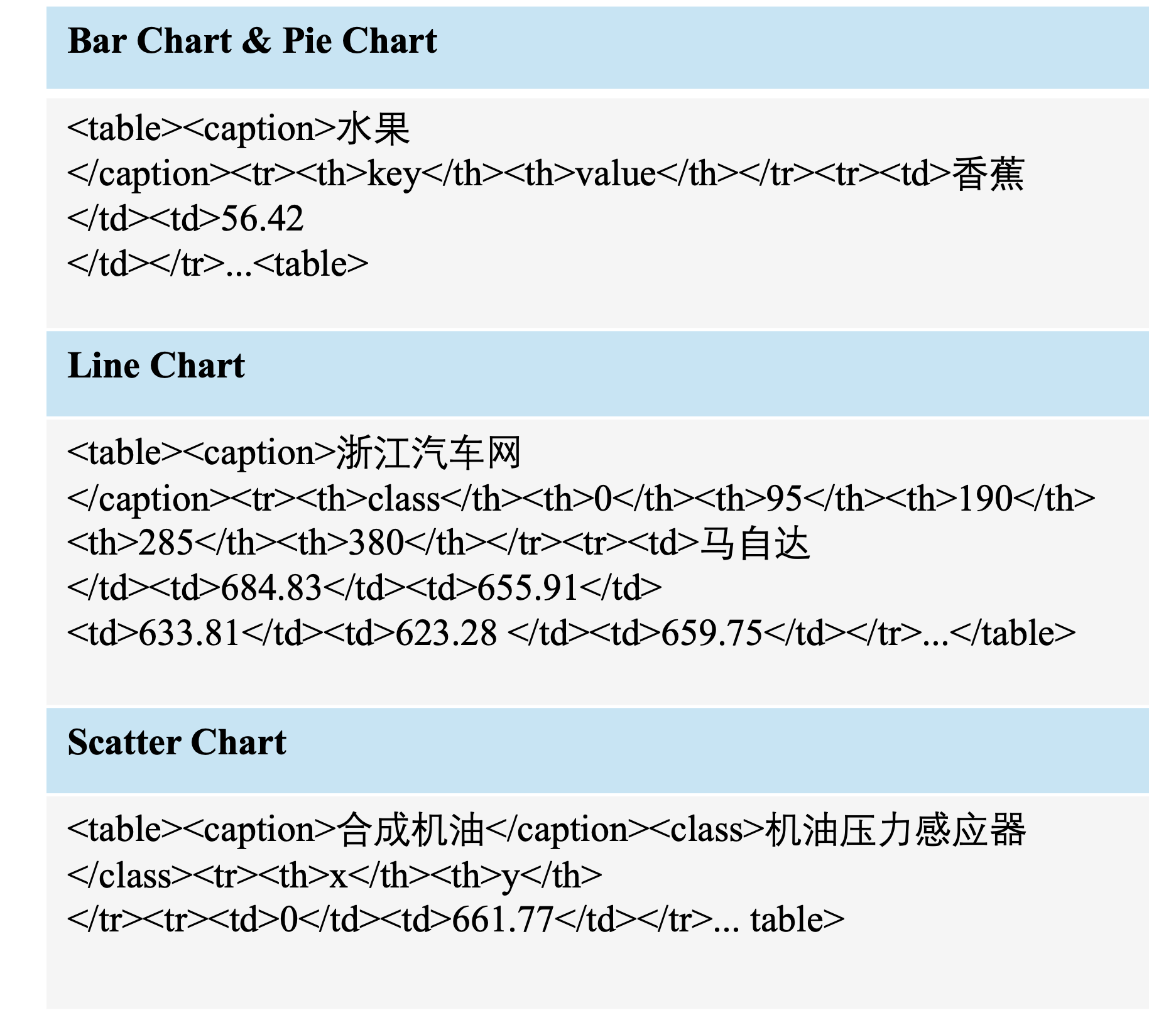}
    \label{fig:groundtruth_table}
    \end{minipage}
  }
  \caption{
    (a) Samples of the synthesized charts: Pie Chart, Vertical Bar Chart, Scatter Chart and Line Chart. (b) The annotation formats corresponding to different charts, which are presented in HTML format.
  }
  \label{fig:charts}
\end{figure*}

\subsubsection{Region Controller.}
The Region Controller plays a pivotal role in the document's structural integrity by meticulously segmenting the data areas into distinct regions for text, images, tables, and charts. It operates on a macro level, determining where each type of content will be placed to optimize readability and visual impact. This controller ensures that the document's layout supports a logical flow, with areas designated for complex data representations such as charts and tables, and separate sections for textual content. By carefully allocating space for each element, the Region Controller ensures that the document's overall composition is balanced and adheres to the principles of good document design, allowing readers to navigate the information with ease.

\subsubsection{Line Controller.}
The Line Controller is responsible for the micro-level organization of textual content within the document. It takes the individual word images produced by the Text Renderer and arranges them into coherent lines, respecting the predefined attributes such as word spacing, line height, and alignment. This controller's work is crucial for establishing the document's typographical style, which includes setting the rhythm and pacing of the text. By fine-tuning the line breaks, indentations, and other typographical elements, the Line Controller ensures that the text is not only legible but also visually engaging. This attention to detail in formatting contributes to a professional and polished appearance, enhancing the document's overall presentation quality.


\subsection{Content Rendering}
With the layout meticulously established, the pipeline transitions to the content rendering phase, where the visual and textual elements of the document come to life. This stage involves the intricate process of integrating graphics and text, ensuring that each component not only complements the layout but also enhances the document's overall narrative and aesthetic appeal.

\begin{figure}[tb]
  \centering
  \includegraphics[width=0.5\textwidth]{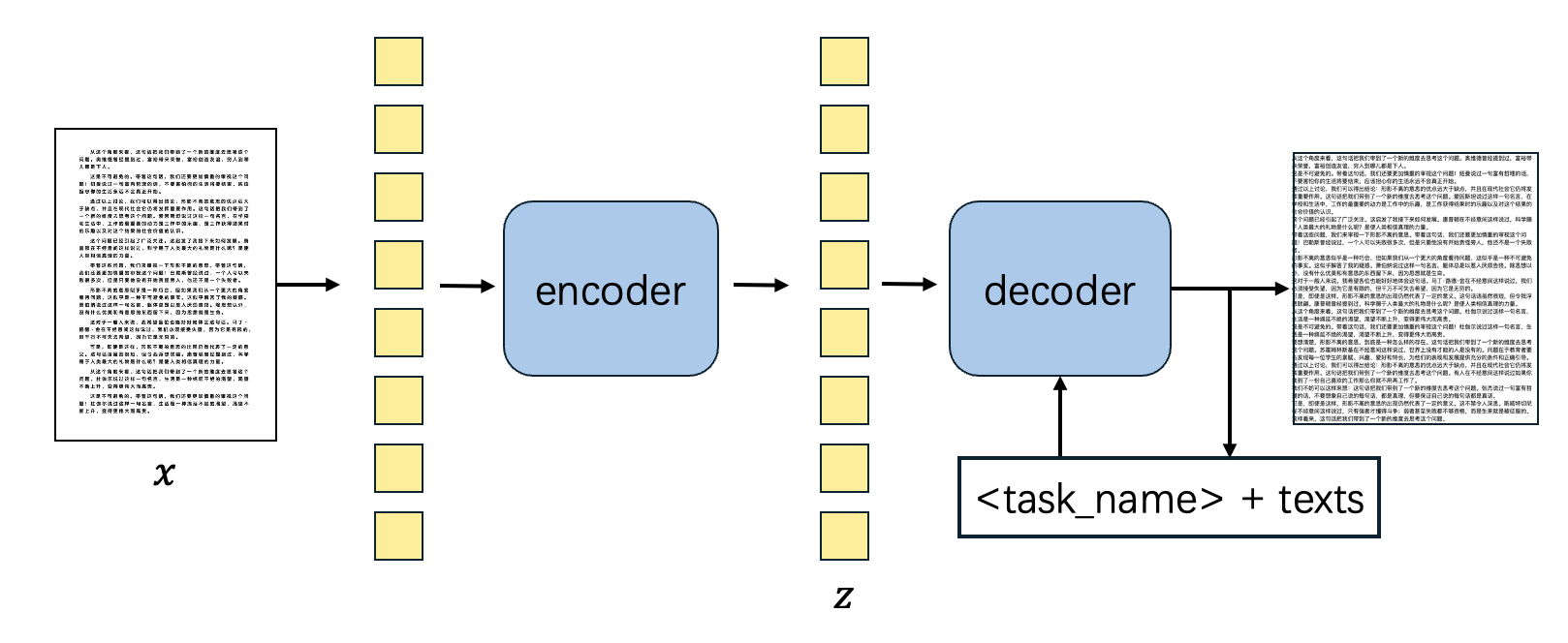}
  \caption{
    This is an overview architecture to training the model
  }
  \label{fig:architecture}
\end{figure}

\subsection{Experimental Results}
\label{sec:results}

\subsubsection{Graphic Renderer.}
The Graphic Renderer is a sophisticated component of our pipeline, dedicated to the rendering of images, tables, and charts. For images, we focus on incorporating natural images, where available category data is used to caption and embed the images within the document. If category information is present, the returned text represents the category; otherwise, it is replaced with a generic placeholder "<nature\_image>". This approach ensures that each image is contextually relevant and enhances the document's informational content.

In the realm of tables, we have designed two distinct types to accommodate various data presentations. The first type features complete borders, suitable for complex data with line breaks within cells, while the second type adopts a minimalist or borderless style, aligning with the prevalent aesthetic in research publications. Both types incorporate random cell merging to manage data complexity effectively. The rendered tables are displayed in Fig.~\ref{fig:table}.

For charts, our pipeline supports the rendering of four chart types: bar, pie, line, and scatter plots. Bar charts, available in both horizontal and vertical orientations, are crafted for data comparison, with key-value pairs represented in a tabular format to facilitate readability. To mitigate issues with overlapping labels in vertical bar charts, we implement random fonts and rotation angles. Pie charts, similar in rendering to bar charts, require that the aggregated values represent a total of 1 or 100, expressed as decimals or percentages. Line charts illustrate trends over time or variables, with each chart featuring a unique set of data groups and points, generating an image-label pair. Scatter plots, used to depict the distribution of a single element, employ a label and x and y coordinates for each point, with the number of points limited to a range of [5, 20] to manage complexity. The generated examples are depicted in Fig.~\ref{fig:element4}. The corresponding HTML annotations are displayed in Fig.~\ref{fig:groundtruth_table}.

The emphasis on the model's ability to understand the structure of diverse elements is paramount. We refrain from using AI tools to generate data within elements, instead leveraging an open textual corpus for our tables and charts, ensuring the authenticity and relevance of the data. The matplotlib library is utilized for chart rendering, and we have refined table rendering techniques to better integrate with the document's overall design.

\begin{table*}[tb]
  \caption{The comparison between different methods across diverse synthetic documents.
  }
  \label{tab:OCR results}
  \centering\setlength{\tabcolsep}{10pt}
  \begin{tabular}{@{}ll|cc|ccc|c@{}}
    \toprule
    \multirow{2}{*}{Metrics} & \multirow{2}{*}{Methods} & \multicolumn{2}{c|}{Pure Document} & \multicolumn{3}{c|}{Complex Document} & \multirow{2}{*}{Average} \\
    \cmidrule{3-7}
    & & {English} & {Chinese} & {Doc w/image} & {Doc w/table} & {Doc w/chart} \\
    \midrule
    \multirow{3}{*}{AED $\downarrow$ }  & Donut~\cite{donut} & 0.3764 & 0.5148 & 0.7631 & 0.8679 & 0.9097 & 0.6864 \\
         & vary~\cite{wei2023vary}       & 0.1452 & 0.1760 & 0.5598 & 0.7415 & 0.6663 & 0.4578 \\
         & our        & \textbf{0.0321} & \textbf{0.1370} & \textbf{0.1665} & \textbf{0.0583} & \textbf{0.1029} & \textbf{0.0994} \\
    \midrule
    \multirow{3}{*}{F1-score $\uparrow$ }  & Donut~\cite{donut} & 0.9370 & 0.8107 & 0.3720 & 0.4573 & 0.2840 & 0.5722 \\
         & vary~\cite{wei2023vary}       & 0.8554 & 0.9002 & 0.5852 & 0.5854 & 0.6531 & 0.7159 \\
         & our        & \textbf{0.9611} & \textbf{0.9020} & \textbf{0.8855} & \textbf{0.9199} & \textbf{0.8810} & \textbf{0.9099} \\
    \midrule
    \multirow{3}{*}{Prediction $\uparrow$ }  & Donut~\cite{donut} & 0.9534 & 0.8256 & 0.4061 & 0.5302 & 0.4063 & 0.6243 \\
         & vary~\cite{wei2023vary}       & 0.8762 & 0.8974 & 0.6383 & 0.7026 & 0.7961 & 0.7821 \\
         & our        & \textbf{0.9717} & \textbf{0.9136} & \textbf{0.9065} & \textbf{0.9347} & \textbf{0.9017} & \textbf{0.9256} \\
    \midrule
    \multirow{3}{*}{Recall $\uparrow$ }  & Donut~\cite{donut} & 0.9228 & 0.8015 & 0.3647 & 0.4313 & 0.2540 & 0.5549 \\
         & vary~\cite{wei2023vary}       & 0.8482 & \textbf{0.9044} & 0.5746 & 0.5501 & 0.5868 & 0.6928 \\
         & our        & \textbf{0.9515} & 0.8916 & \textbf{0.8682} & \textbf{0.9076} & \textbf{0.8636} & \textbf{0.8965} \\
  \bottomrule
  \end{tabular}
\end{table*}

\subsubsection{Text Renderer.}
The Text Renderer plays an indispensable role in the content rendering process, meticulously generating word images for each word in the text. This method affords a high level of control over the typography and layout, ensuring that the text is not only legible but also aesthetically integrated with the document's visual elements. The Text Renderer works in concert with the Graphic Renderer to weave a cohesive and engaging narrative, blending visual and textual information to enhance the reader's experience.

Following Donut's data generation approach, the Text Renderer creates a word image for each word, which is crucial for the document's visual composition and label generation. This attention to detail in text rendering ensures that the document's textual content is as carefully crafted as its visual elements, contributing to a polished and professional final product.

\begin{table}[tb]
  \caption{Performance Comparison of different methods on CORD.}
  \label{tab:cord-results}
  \centering\setlength{\tabcolsep}{4pt}
  \begin{tabular}{@{}lc|cccc@{}}
    \toprule
    Model & OCR & Acc & Precision & Recall & F1 \\
    \midrule
    BERT~\cite{hwang2019post} & $\surd$ & 78.2 & - & - & 82.2 \\
    BROS~\cite{hong2022bros} & $\surd$ & 80.3 & - & - & 83.7 \\
    LayoutLMv2~\cite{xu2020layoutlmv2} & $\surd$ & 87.0 & - & - & 88.9 \\
    KOSMOS-2.5~\cite{lv2023kosmos} & - & - & 83.64 & 87.83 & 85.69 \\
    Donut~\cite{donut} & - & 93.5 & - & - & 91.6 \\
    \midrule
    our & - & 90.1 & 82.6 & 83.3 & 82.9 \\
  \bottomrule
  \end{tabular}
\end{table}

\subsection{Concerns of Data Generation Pipeline}
\subsubsection{Scalability}
Even if we generate as much diverse data as possible, it hardly covers all real-world document layouts. To mitigate this, we've integrated real document images into our benchmark to maximize layout variability.
However, it is worth noting that our solution is highly adaptable, with scalability in two key dimensions: 1) \textbf{Layout Customization:} We allow for tailored document layouts to swiftly and cost-effectively expand our training data to fit various scenarios. 2) \textbf{Language Independence:} Our pipeline transcends language barriers, enabling document image generation in any language.  For instance, we've produced French documents using the ROOTS\cite{laurenccon2022bigscience} dataset.

\begin{figure*}[tb]
  \centering
  \subfloat[ ]{
    \begin{minipage}[b]{0.33\linewidth}
    \centering
    \includegraphics[height=2.2in]{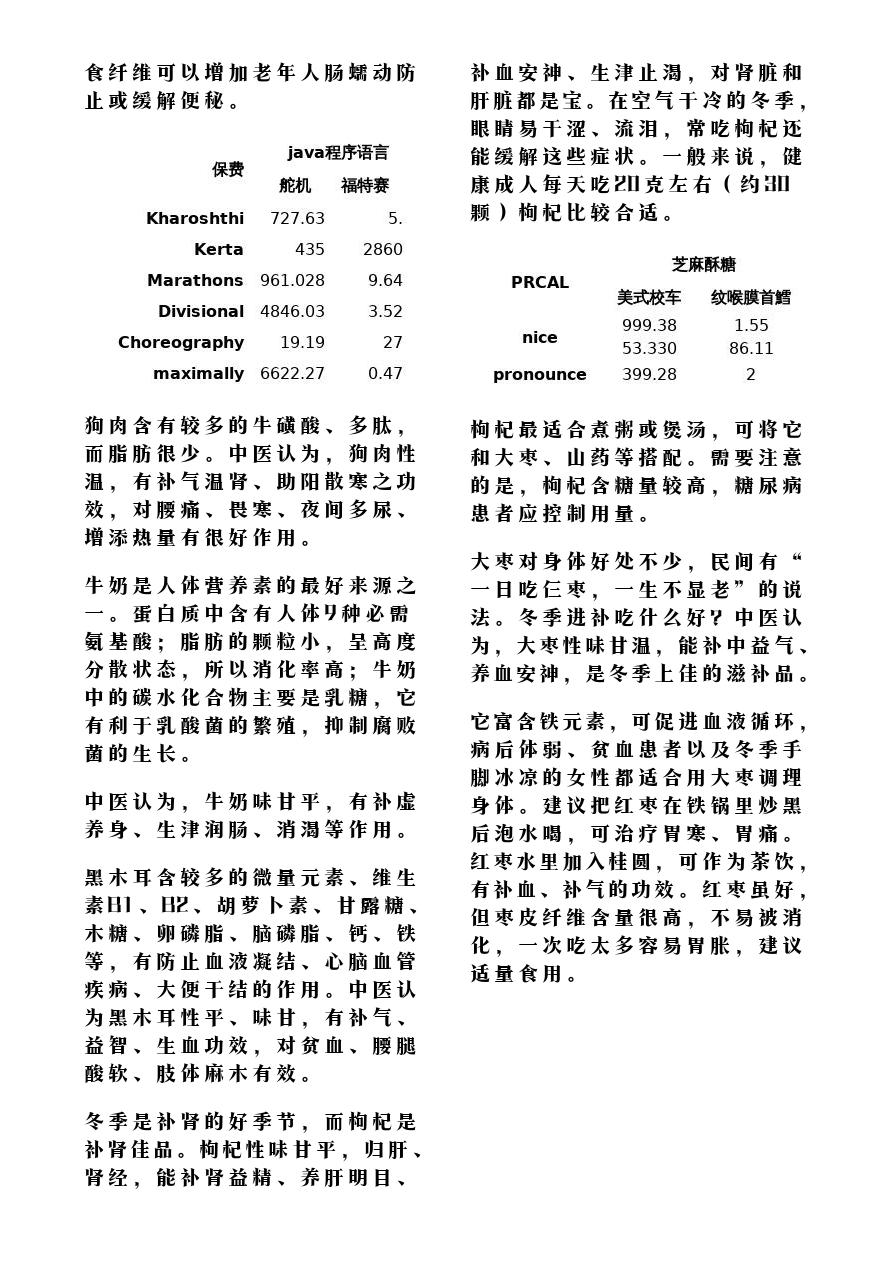}
    \label{fig:tableDoc}
    \end{minipage}
  }
  \subfloat[ ]{
    \begin{minipage}[b]{0.33\linewidth}
    \centering
    \includegraphics[height=2.2in]{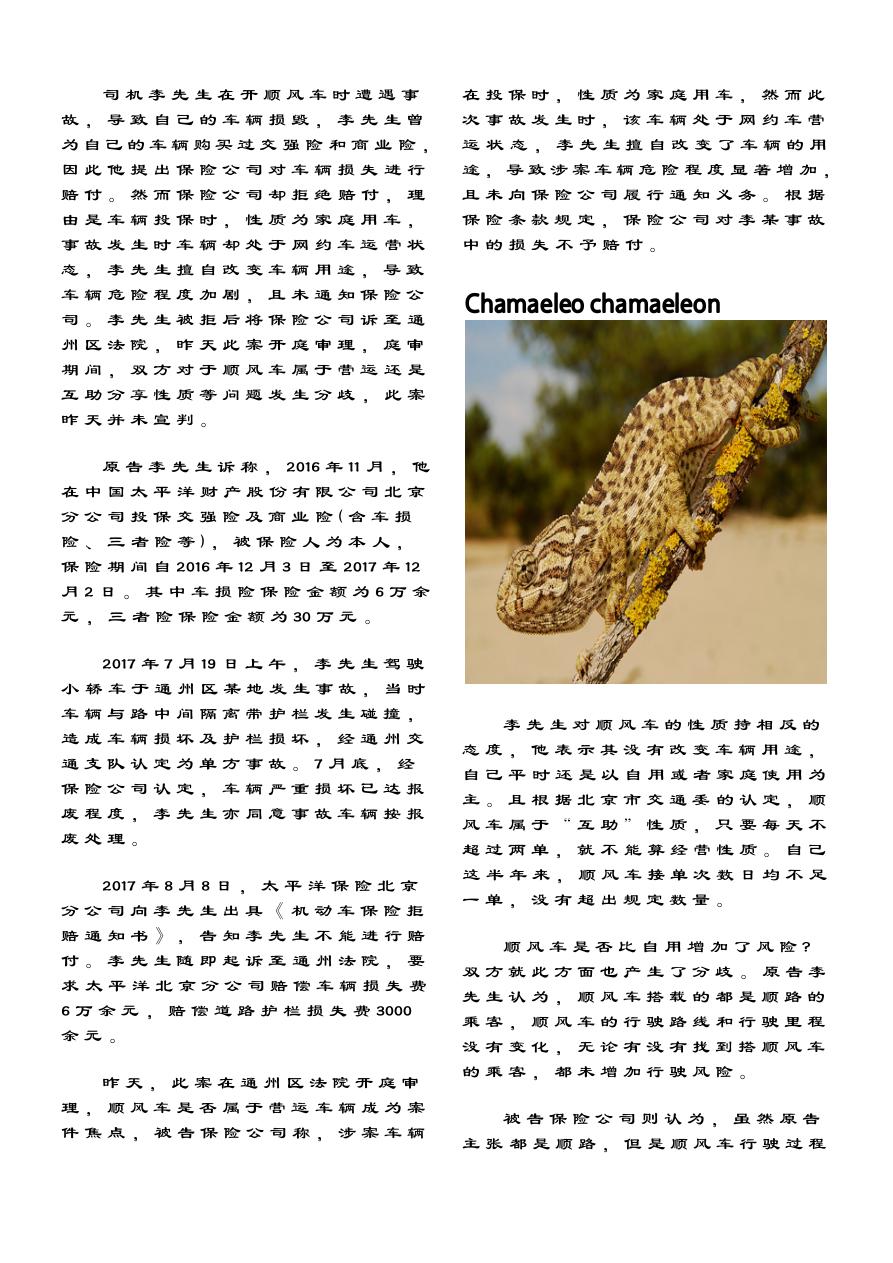}
    \label{fig:imageDoc}
    \end{minipage}
  }
  \subfloat[ ]{
    \begin{minipage}[b]{0.33\linewidth}
    \centering
    \includegraphics[height=2.2in]{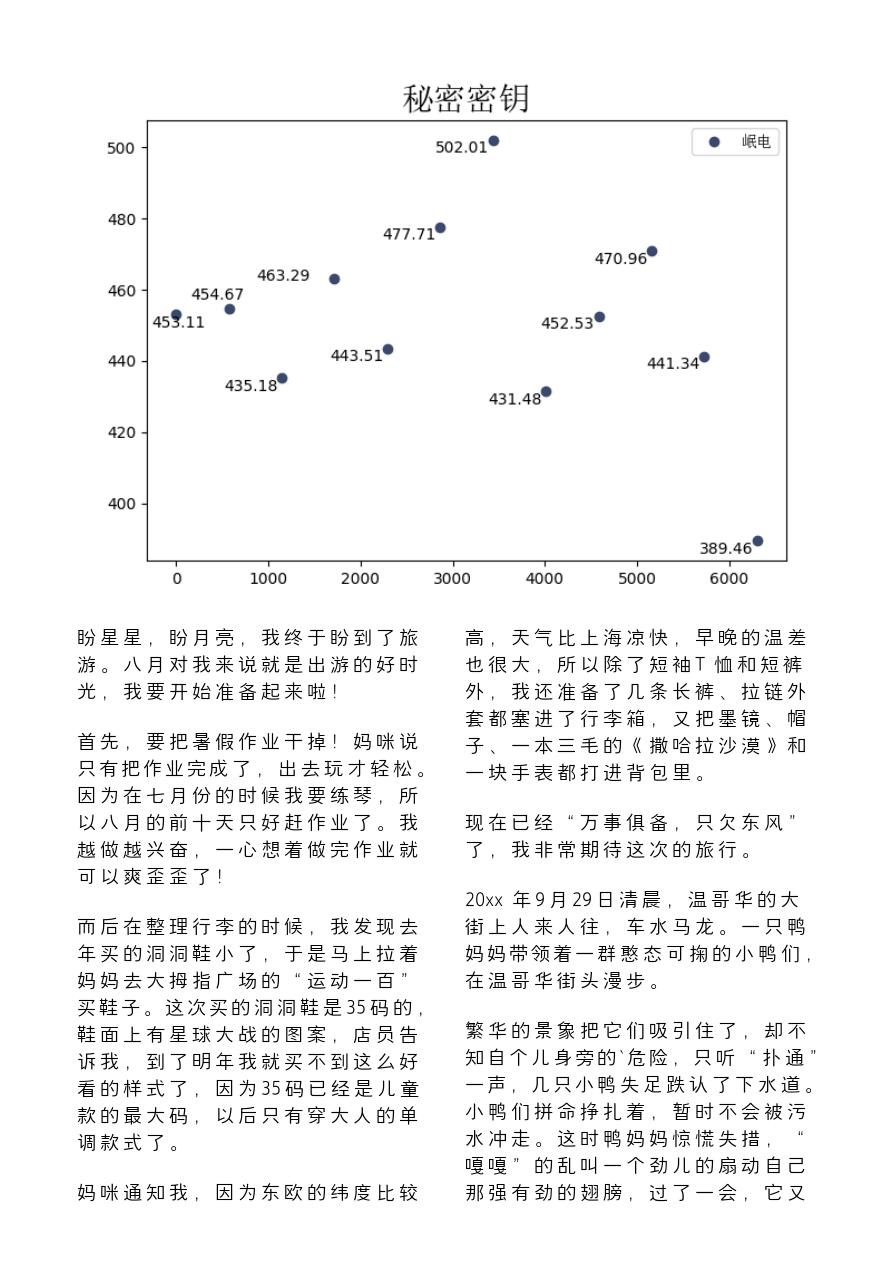}
    \label{fig:scatterDoc}
    \end{minipage}
  }
  \quad
  \subfloat[ ]{
    \begin{minipage}[b]{0.33\linewidth}
    \centering
    \includegraphics[height=2.2in]{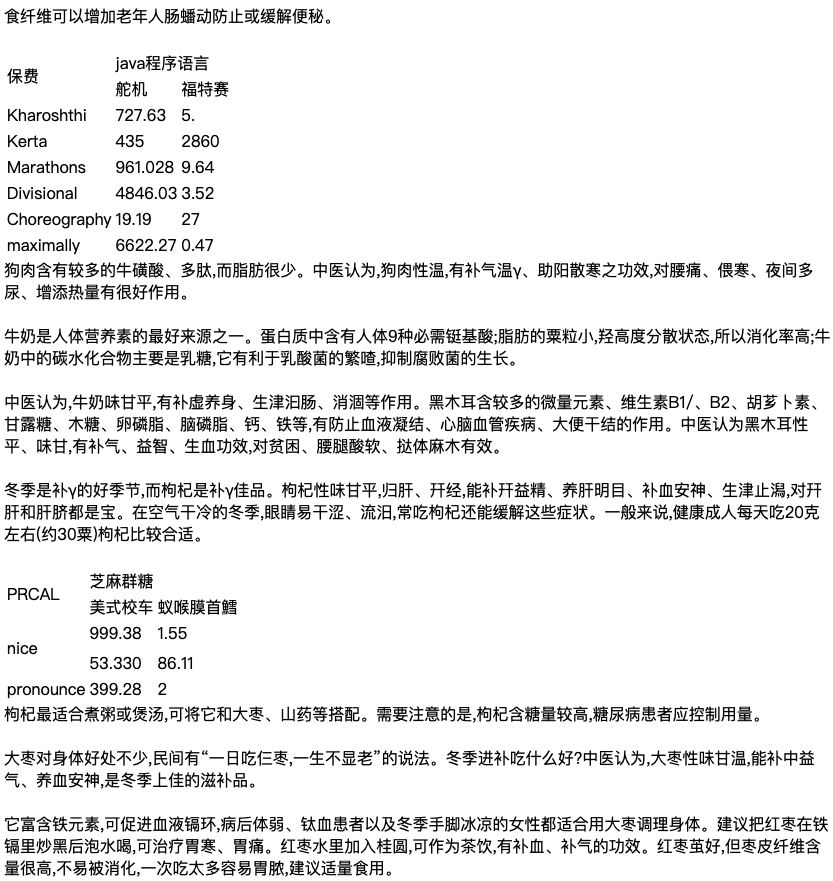}
    \label{fig:tableDoc_gt}
    \end{minipage}
  }
  \subfloat[ ]{
    \begin{minipage}[b]{0.33\linewidth}
    \centering
    \includegraphics[height=2.2in]{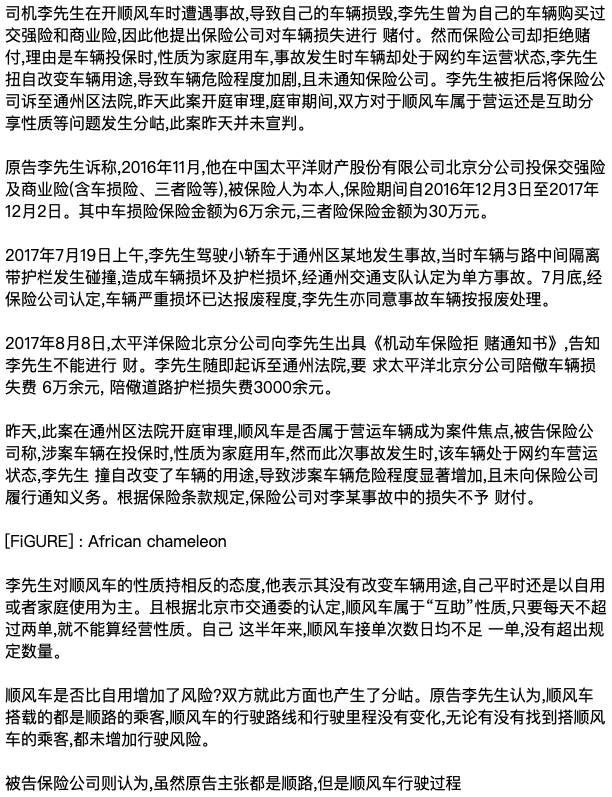}
    \label{fig:imageDoc_gt}
    \end{minipage}
  }
  \subfloat[ ]{
    \begin{minipage}[b]{0.33\linewidth}
    \centering
    \includegraphics[height=2.2in]{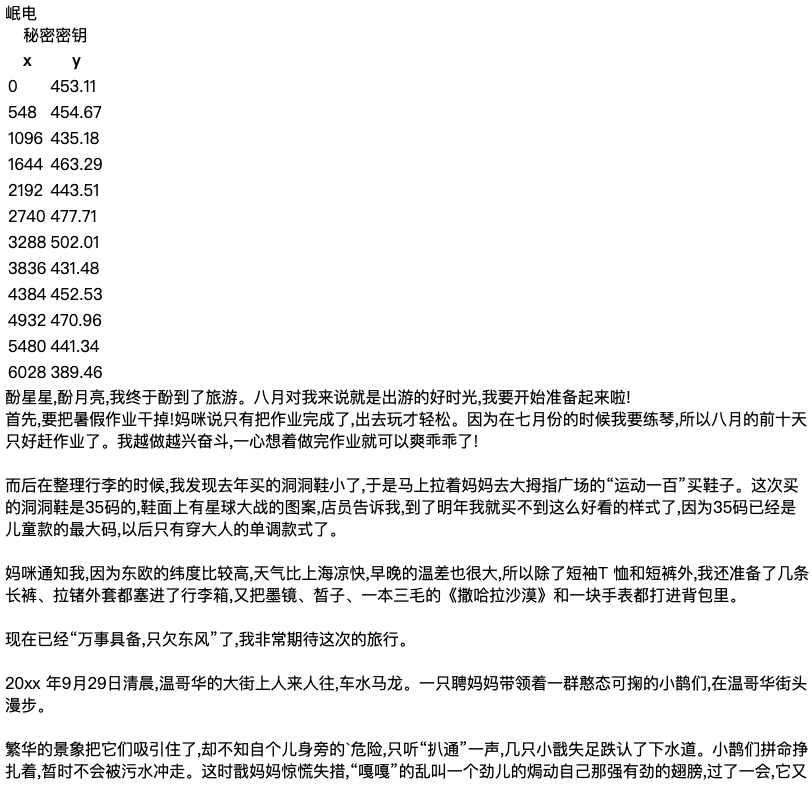}
    \label{fig:scatterDoc_gt}
    \end{minipage}
  }
  \caption{Examples of document image parsing on synthesized document with tables, images, and chart. (a), (b) and (c) stand for the synthetic document images with tables, images, and chart, (d), (e), and (f) represent the parsing results of the model on them, respectively.}
  \label{fig:doc-table_img}
\end{figure*}

\subsubsection{Data Privacy}
Our pipeline allows for local regulatory adaptation and reproducibility of datasets through customizable pipeline components. We advocate for the use of public corpora and tools to foster transparency and verifiability in research.

\section{Training on SynthDoc}
\label{sec:training}
This section details the pre-training of the model based on the Donut architecture, focusing on its parsing performance with bilingual (English and Chinese) documents. The primary objective is to validate the model's ability to effectively handle and interpret content in both languages, ensuring its suitability for multilingual document analysis.

\subsection{Model Architecture}
\label{sec:model}
Unlike previous OCR-based approaches~\cite{huang2022layoutlmv3,bai2022wukong} for visual document understanding tasks, recent research~\cite{Pix2struct,dhouib2023docparser} has shifted towards parsing document images in an end-to-end fashion, eliminating the need for OCR results as input. The dataset we generated primarily aims to enhance and validate the visual document parsing capabilities of this end-to-end models. 
Illustrated in \Cref{fig:architecture}, our model is constructed based on the Donut architecture. We follow the Donut~\cite{donut}, utilizing the Swin-Transformer~\cite{liu2021swin} as our visual encoder. Previous experiments have demonstrated its superior performance compared to ViT~\cite{dosovitskiy2020image}. We employ mBART~\cite{lewis2019bart} as the decoder, which has stronger noise robustness and multilingual capabilities. 

\subsection{Implementation Details}
\label{sec:setups}
Following the previous works~\cite{blecher2023nougat,donut}, we employ Swin-Base as the encoder and the first four layers of mBART as the decoder, with a patch size of 4 and a window size of 10. We set the input image size to (H, W) = (1280, 960) to meet the requirements of Swin-Base for image dimensions. For pre-training, we set a batch size of 192 and employ the AdamW optimizer, initializing the learning rate at 5e-5 and setting a minimum of 7.6e-6, while utilizing an exponential scheduler with a gamma of 0.9996, updating the learning rate every 16 training steps. For fine-tuning, we utilize a cosine scheduler with a learning rate of 3e-5 to optimize our model, dynamically adjusting the input size according to the datasets, a practice effectively demonstrated by Donut.

\begin{figure*}[tb]
  \centering
  \subfloat[ ]{
    \begin{minipage}[b]{0.25\linewidth}
    \centering
    \includegraphics[height=2.1in]{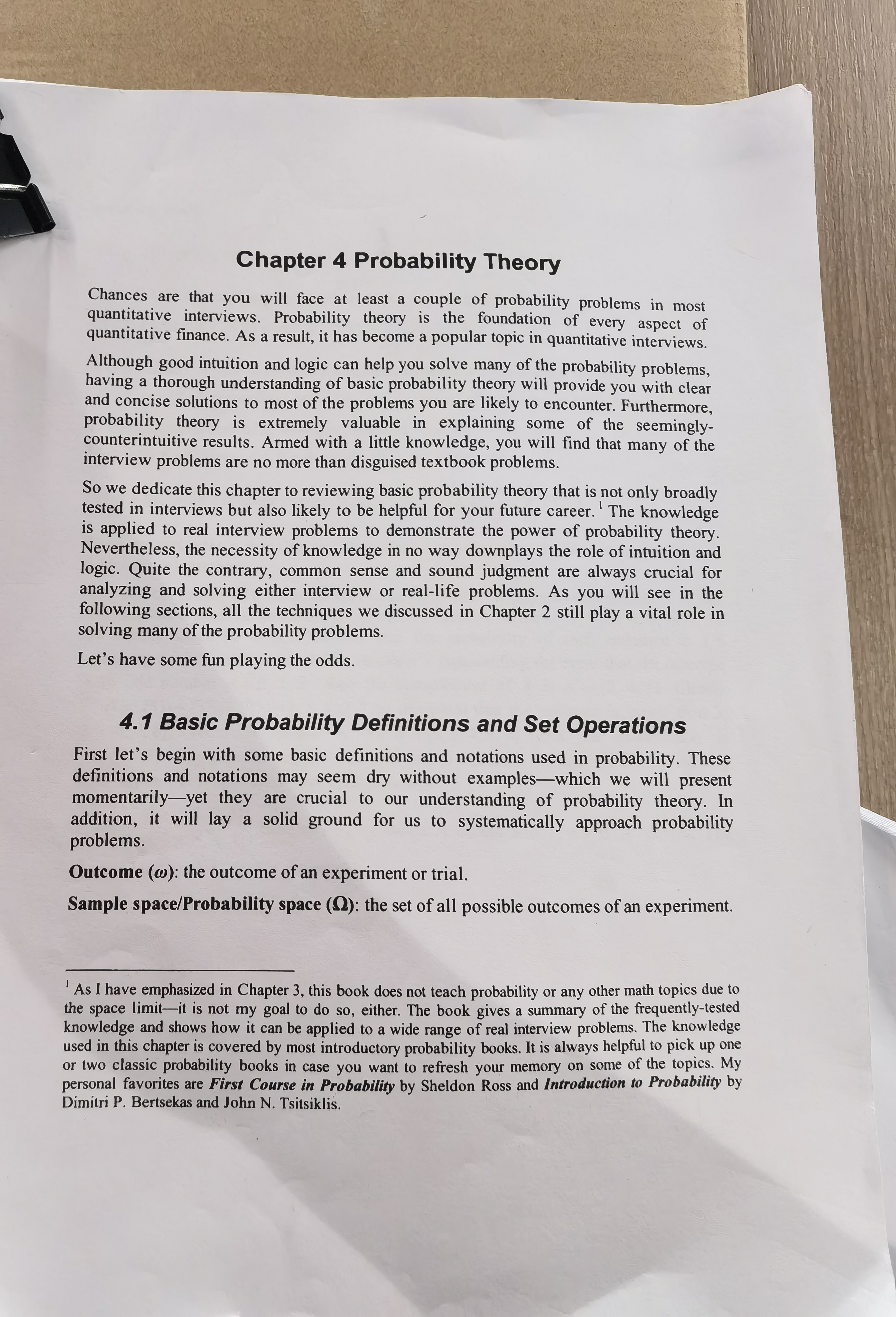}
    \label{fig:demo1}
    \end{minipage}
  }
  \subfloat[ ]{
    \begin{minipage}[b]{0.25\linewidth}
    \centering
    \includegraphics[height=2.1in]{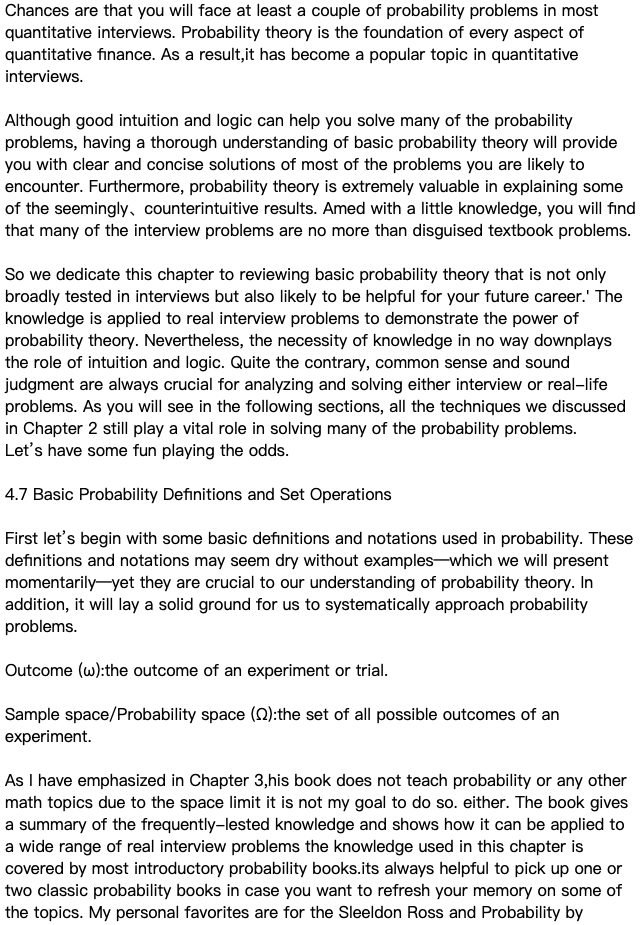}
    \label{fig:demo1_gt}
    \end{minipage}
  }
  \subfloat[ ]{
    \begin{minipage}[b]{0.25\linewidth}
    \centering
    \includegraphics[height=2.1in]{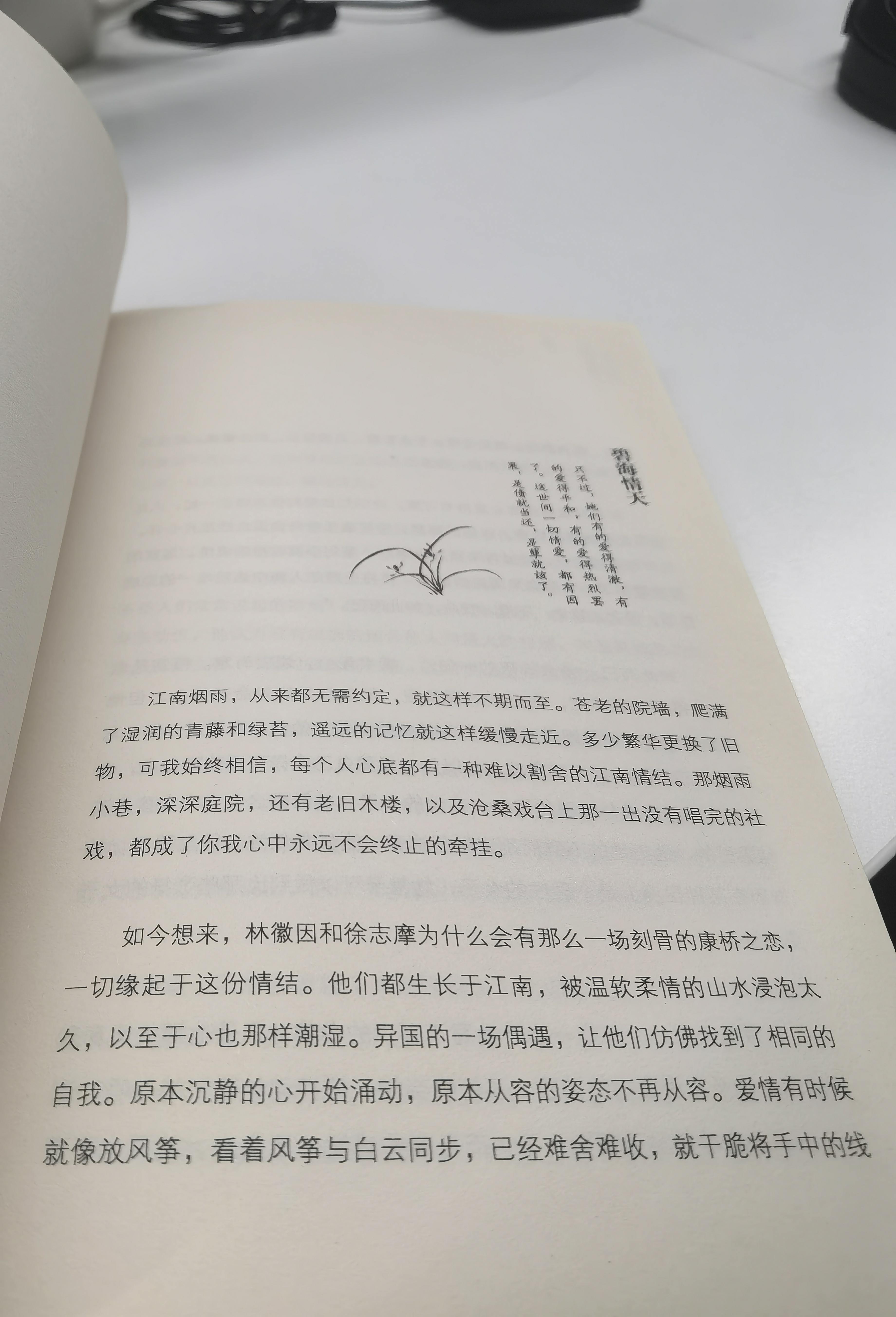}
    \label{fig:demo3}
    \end{minipage}
  }
  \subfloat[ ]{
    \begin{minipage}[b]{0.25\linewidth}
    \centering
    \includegraphics[height=2.1in]{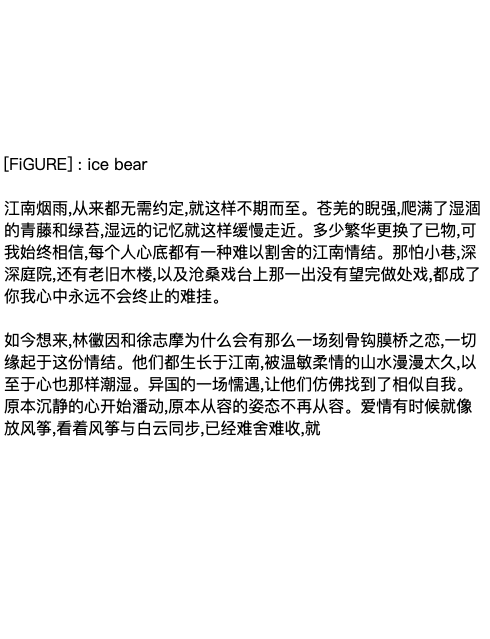}
    \label{fig:demo3_gt}
    \end{minipage}
  }
  \caption{Examples of document image parsing on real English and Chinese documents. (a) real English document (b) prediction of English documents (c) real Chinese document (d) prediction of Chinese documents.}
  \label{fig:real-doc_img}
\end{figure*}

\subsubsection{Document Image Parsering.}
\label{sec:parsering}
We evaluate the document parsing capabilities of other end-to-end models using the proposed benchmark in this paper and compare them with the performance of the model we trained. As shown in \Cref{tab:OCR results}, we evaluate the models on five types of documents: English documents, Chinese documents, documents with natural images, documents with tables, and documents with charts. We observed that all models exhibited strong performance on English and Chinese documents, except for Donut, which showed slightly inferior results on the Average Edit Distance (AED), possibly due to its lack of training on the document dataset. However, with the exception of our model, all models displayed inadequate performance on complex documents containing additional elements. Specifically, our model achieved 0.1665, 0.0583, and 0.1029 AED on document images with images, tables, and charts, respectively, showing reductions of 0.3933, 0.6832, and 0.5634 compared to the Vary. It is noteworthy that in our benchmark, text labels associated with other elements represent only a small portion. This observation indicates that elements such as images in documents can significantly impact the model's text parsing capability.

\subsubsection{Results on CORD.}
\label{sec:results_cord}
The CORD dataset is a collection of data used for receipt recognition, comprising 800 samples for training and 100 samples for testing. Our pipeline's performance on the English CORD dataset did not demonstrate the expected improvements, due to the substantial distribution bias towards Chinese, which can be addressed by enhancing our model to more adeptly handle English-language documents in subsequent research. However, it is worth noting that our model not only improves its proficiency in Chinese document image recognition but also ensures comparable performance in downstream tasks.

\subsection{Visual Analysis}
\label{sec:visual_analysis}
We provide sufficient visualization results of our model to demonstrate the excellent performance of the model in text image recognition.  Specifically, The ~\Cref{fig:doc-table_img} illustrates synthetic images, containing tables, images, and charts demonstrating our model's ability to parse text, tables, images, and charts information in  a manner consistent with human reading order. Furthermore, as illustrated in the last row of ~\Cref{fig:real-doc_img}, our model exhibits robust parsing capability when applied to real document images.

\subsubsection{Spatial Understanding.}
We observed that end-to-end models possess strong spatial understanding capabilities. Specifically, we provided serialized numerical coordinates in scatter plots and line graphs, defining a new coordinate space. Our trained model can accurately identify the localization of points in this coordinate space. As shown in ~\Cref{fig:scatterDoc} is the document image with a scatter chart, and ~\Cref{fig:scatterDoc_gt} is the model's prediction, we only provided the vertical coordinates of the points in the image. However, the model can accurately identify their corresponding horizontal coordinates. For example, for a point with a vertical coordinate of 435.18, the model can identify its horizontal coordinate as 1096, which closely aligns with our provided ground truth. 

\subsubsection{Robust Interference Capability.}
Benefiting from training our model with documents containing natural images, our model exhibits robust interference capability. As shown in ~\Cref{fig:real-doc_img}, ~\Cref{fig:demo1} and ~\Cref{fig:demo3} presents a real image captured by a camera, while ~\Cref{fig:demo1_gt} and ~\Cref{fig:demo3_gt} illustrates the model's prediction. Despite incorrectly identifying some challenging regions as natural images, it does not impede subsequent text parsing. This phenomenon has not been observed in other end-to-end methods. We believe that training with synthetic data incorporating various contexts is an important approach to improving model robustness and performance.

\section{Limitation}
\label{sec:limitation}
While the current generation of documents through SynthDoc is a significant step forward, we acknowledge that the types of documents created thus far are somewhat limited in variety. To enhance the richness of our dataset and to better mimic the complexity of real-world documents, we are committed to expanding our pipeline's capabilities. Future iterations will incorporate more sophisticated intermingling of document elements, allowing for the generation of even more intricate and varied document types. This evolution will not only challenge and refine existing models but also pave the way for the development of more advanced document image recognition systems, capable of handling the multifaceted nature of documents encountered in everyday applications.

\section{Conclusion}
\label{sec:conclusion}
In conclusion, this study presents SynthDoc, an innovative pipeline for generating synthetic documents, which plays a pivotal role in bolstering Visual Document Understanding (VDU). By producing a high-quality, diverse dataset that encompasses text, images, tables, and charts, SynthDoc addresses the critical issues of data acquisition and the constraints imposed by current datasets. Utilizing publicly accessible corpora and sophisticated rendering tools, SynthDoc has successfully created a dataset that is both extensive and adaptable. Our empirical evaluations, employing the Donut model, have shown that models trained on SynthDoc's dataset not only excel in pre-training read tasks but also exhibit resilience in downstream tasks, even when faced with linguistic disparities. The introduction of a benchmark dataset featuring 5,000 image-text pairs not only highlights the capabilities of our pipeline but also serves as a substantial contribution to the VDU community, facilitating further research and development in the realm of document image recognition. This research marks a significant advancement in the field by providing a scalable approach to overcoming data scarcity and by empirically validating the effectiveness of end-to-end models in parsing intricate, real-world documents.

\bibliographystyle{ACM-Reference-Format}
\bibliography{camera-ready}










\end{document}